\documentclass{article}

\usepackage{PRIMEarxiv}

\usepackage{amsmath}

\usepackage{caption}
\usepackage{wrapstuff}

\usepackage[table]{xcolor}
\usepackage{booktabs}
\usepackage{wrapfig}
\definecolor{methodRow}{RGB}{214,226,245}

\usepackage{subcaption}
\usepackage{colortbl}
\usepackage{tabularx}
\usepackage{booktabs}
\usepackage{listings}
\usepackage{xcolor}

\definecolor{codebg}{RGB}{246,247,249}
\definecolor{codeframe}{RGB}{205,208,213}
\definecolor{cfgkey}{RGB}{0,92,160}
\definecolor{cfgstr}{RGB}{178,76,0}

\lstdefinestyle{cfgcompact}{
  backgroundcolor=\color{codebg},
  basicstyle=\ttfamily\scriptsize,
  keywordstyle=\color{cfgkey}\bfseries,
  stringstyle=\color{cfgstr},
  numberstyle=\tiny\color{gray},
  numbers=left,
  numbersep=4pt,
  frame=single,
  rulecolor=\color{codeframe},
  breaklines=true,
  breakatwhitespace=false,
  columns=fullflexible,
  keepspaces=true,
  tabsize=2,
  showstringspaces=false,
  captionpos=b,
  xleftmargin=6pt,
  xrightmargin=6pt,
  aboveskip=4pt,
  belowskip=4pt
}

\usepackage[utf8]{inputenc} 
\usepackage[T1]{fontenc}    
\usepackage{hyperref}       
\usepackage{url}            
\usepackage{booktabs}       
\usepackage{amsfonts}       
\usepackage{nicefrac}       
\usepackage{microtype}      
\usepackage{lipsum}
\usepackage{fancyhdr}       
\usepackage{graphicx}       
\graphicspath{{media/}}     

\title{TEEM: Task-Guided Event Evidence Memory for Long-Horizon Manipulation
}

\author{
  Bang Hu \\
  Fudan University \\
   \And
  Jiacheng Liu \\
  Westlake University \\
     \And
  Ping Chen, Fan Zhang, Wei Cao \\
    Fudan University \\
}

\begin{document}
\maketitle

\begin{abstract}
Memory remains a challenge in long-horizon robotic manipulation, particularly for VLA policies that must rely on visual information that may disappear before a later action requires it. 
We introduce \textbf{TEEM} (\emph{Task-Guided Event Evidence Memory}), which uses subtask-linked targets to guide the sampling of task-relevant visual evidence.
At initialization, a VLM planner organizes the instruction into ordered, subtask-linked targets and evidence descriptions. 
During execution, we use motion prominence $P$ to identify observations near deceleration-to-stability transitions,
target-conditioned visual quality $Q$ evaluates the active target regions, 
and temporal stability $S$ selects a representative RGB observation for each plan node. 
TEEM supplies these observations through the receiving policy's native memory pathway without changing its backbone or action-generation mechanism.
We evaluate TEEM with MemoryVLA on LIBERO-Long, with EventVLA on RoboTwin-MeM, and on three physical-robot tasks.
On LIBERO-Long, TEEM matches or improves success on every evaluated task, raising mean success from 90.0\% to 91.8\% on Long-10 and from 93.4\% to 94.2\% on the selected Long-90 subset. 
On RoboTwin-MeM, TEEM reaches an 86.0\% mean success rate, compared with 77.1\% for EventVLA under the same evaluation. 
On a DOBOT CR5V, TEEM also achieves remarkable improvement with the 5$\pi_{0.5}$ policy.
These results show that TEEM can improve access to earlier visual evidence across distinct VLA memory pathways and physical-robot execution.
\end{abstract}

\section{Introduction}
\label{sec:introduction}

Vision-language-action (VLA) models have made remarkable progress in robotic manipulation, mapping visual observations and language instructions to increasingly general-purpose robot behaviors~\cite{brohan2023rt2,kim2024openvla,black2024pi0}. 
 However, long-horizon manipulation remains challenging because a later action may depend on task-relevant information that is no longer visible in the current scene.
For example, a robot may need to identify an object from a cue briefly revealed beneath a cover, reproduce a demonstrated order of visiting locations, or stop pressing a button after a prescribed number of presses.
We study this non-Markovian setting, in which the current policy input alone may not determine the correct next action: earlier observations and actions can remain necessary to select the appropriate continuation~\cite{shi2025memoryvla,yang2026eventvla}.

Cognitive research offers a useful perspective on long-horizon tasks: activities can be organized into hierarchies of tasks and subtasks~\cite{farooqui2012hierarchical}, and the task being performed influences where people direct their gaze~\cite{rothkopf2007task}.
Together, these findings motivate us to organize evidence acquisition around task structure, so that subtask-linked targets guide which visual evidence is collected and retained throughout execution.
We distinguish a \emph{subtask}, an interaction to be performed, from its \emph{target}, an object or region associated with information that may be needed during task execution. 
This distinction gives rise to two common cases: different subtasks may require evidence from different targets, while distinct events involving the same target may require separate records. 
We refer to using subtask-linked targets to guide evidence acquisition as \emph{task-guided evidence focusing}.

\begin{figure*}[t]
  \centering
  \includegraphics[width=0.85\textwidth]{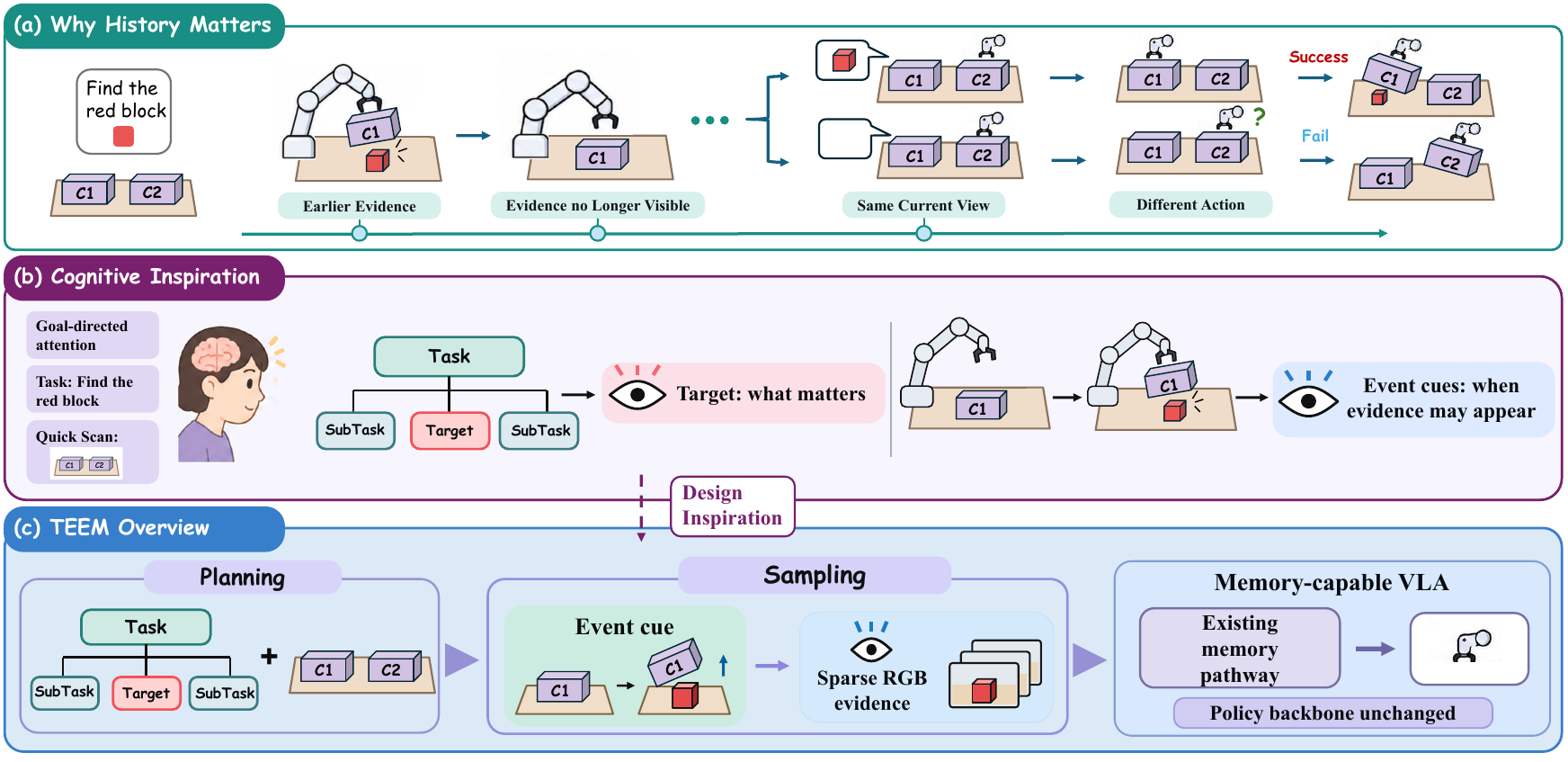}
  \caption{\textbf{TEEM focuses target memory through task-guided evidence focusing.}
(a) Earlier evidence may disappear before a later decision, so identical current observations can call for different actions.
(b) Cognitive findings on hierarchical task organization and task-dependent attention inspire the use of subtask-linked targets to guide evidence acquisition.
(c) TEEM builds an ordered target subtask at initialization, select sparse RGB evidence using motion and visual cues, and supplies to an existing native memory pathway.}
  \label{fig:teem-overview}
\end{figure*}

Building on this perspective, we introduce \textbf{TEEM} (\emph{Task-Guided Event Evidence Memory}), a cognition-inspired external memory module that implements task-guided evidence focusing.
As illustrated in Fig.~\ref{fig:teem-overview}, TEEM comprises \emph{task-guided evidence planning} and \emph{event-driven evidence sampling}, connected through subtask-linked targets.
Given a task-family specification, the instruction, and the initial observation, planning associates subtasks with targets and organizes
the events to be retained.
These targets then guide lightweight evidence sampling during execution, without repeated semantic VLM calls.
Sampling follows two principles: 
\emph{target importance} prioritizes observations relevant to the current subtask's target, using physical event cues to identify candidate
moments and target-region quality and temporal stability to select representative observations;
\emph{target sparsity} consolidates repeated views of the same event into one record while preserving distinct task-relevant occurrences. 
The selected observations are maintained in a fixed-capacity memory bank.
TEEM is designed to integrate with memory-capable VLA policies through their existing memory pathways, without modifying the policy backbone, providing a modular realization of cognition-inspired evidence acquisition and retention.

Our contributions are summarized as follows:
\begin{itemize}
  \item We formulate task-guided evidence focusing for long-horizon manipulation, linking each subtask's memory requirement to a grounded target and planned evidence occurrence. 
  This structure guides targeted keyframe acquisition and lightweight transitions between evidence goals.
  \item We develop TEEM, a pluggable, policy-external module for task-guided evidence sampling. 
  TEEM uses physical event cues to propose candidate windows and selects representative RGB observations based on motion prominence $P$, target-region quality $Q$, and temporal stability $S$, then supplies them through the receiving policy's native memory interface.
  \item We evaluate TEEM with EventVLA on RoboTwin-MeM, MemoryVLA on LIBERO-Long, and $\pi_{0.5}$ on a physical robot. 
  TEEM improves aggregate success over the evaluated native baselines in simulation and real-world. 
  Cross-policy evaluations and ablations characterize its integration scope and the roles of its selection criteria.
\end{itemize}

\section{Related Work}
\label{sec:related-work}

\subsection{VLA Policies for Long-Horizon Manipulation}

Recent VLA policies have made substantial progress in generalizable and language-conditioned robotic manipulation~\cite{brohan2023rt1,brohan2023rt2,embodimentcollaboration2023openx,octomodelteam2024octo,kim2024openvla,black2024pi0,driess2023palme,jiang2023vima,li2024roboflamingo,belkhale2024rth},
yet long-horizon tasks remain difficult because later actions may depend on earlier evidence that is absent from the current observation.
Existing approaches address this temporal dependency in different ways.
One line places a high-level semantic module above the low-level controller.
MemER~\cite{sridhar2025memer}, for example, uses a high-level VLM to select historical observations and produce subtask instructions for a downstream VLA.
Another line compresses interaction history into a representation that can be retrieved or consumed during action generation, as in MemoryVLA~\cite{shi2025memoryvla}, ContextVLA~\cite{jang2025contextvla}, and MemBodied~\cite{pala2026membodied}.
A third line keeps only selected visual events: EventVLA~\cite{yang2026eventvla} predicts future keyframe evidence from policy representations, while KEMO~\cite{zeng2026kemo} stores event-selected visual tokens in temporal order.
TEEM follows this sparse-evidence paradigm and introduces \emph{task-guided evidence focusing}, which uses targets associated with subtasks to organize the memory-acquisition process. 
The sampling module remains external to the action policy and can provide observations through an existing memory pathway without changing the policy backbone or action-generation architecture.

\subsection{Event-Driven Evidence Sampling}

In robotic memory systems, event-driven evidence sampling commonly relies on either physical interaction cues or task-content cues~\cite{yang2026eventvla,zeng2026kemo,smelova2026replanmem,chen2026keyframechaining}. 
Physical-cue methods use kinematics, motion changes, prediction disagreement, or visual change to propose candidate moments.
For example, KEMO~\cite{zeng2026kemo} combines kinematic saliency with visual dissimilarity filtering, while RePlanMem-VLA~\cite{smelova2026replanmem} triggers updates from disagreement between overlapping action predictions and removes redundant observations. 
Task-content methods instead use explicit semantic reasoning or learned task-conditioned prediction: MemER~\cite{sridhar2025memer} nominates historical observations with a high-level VLM, EventVLA~\cite{yang2026eventvla} predicts future keyframe probabilities from policy representations, and Keyframe-Chaining VLA~\cite{chen2026keyframechaining} uses task-modulated transition embeddings and progress-aware retrieval. 
TEEM combines a subtask-linked target description with physical event cues and target-region visual quality, allowing a lightweight external selector to organize evidence acquisition without per-candidate semantic calls or a policy-internal keyframe-prediction head.

\section{Method}
\label{sec:method}

\begin{figure*}[t]
  \centering
  \includegraphics[width=0.9\textwidth]{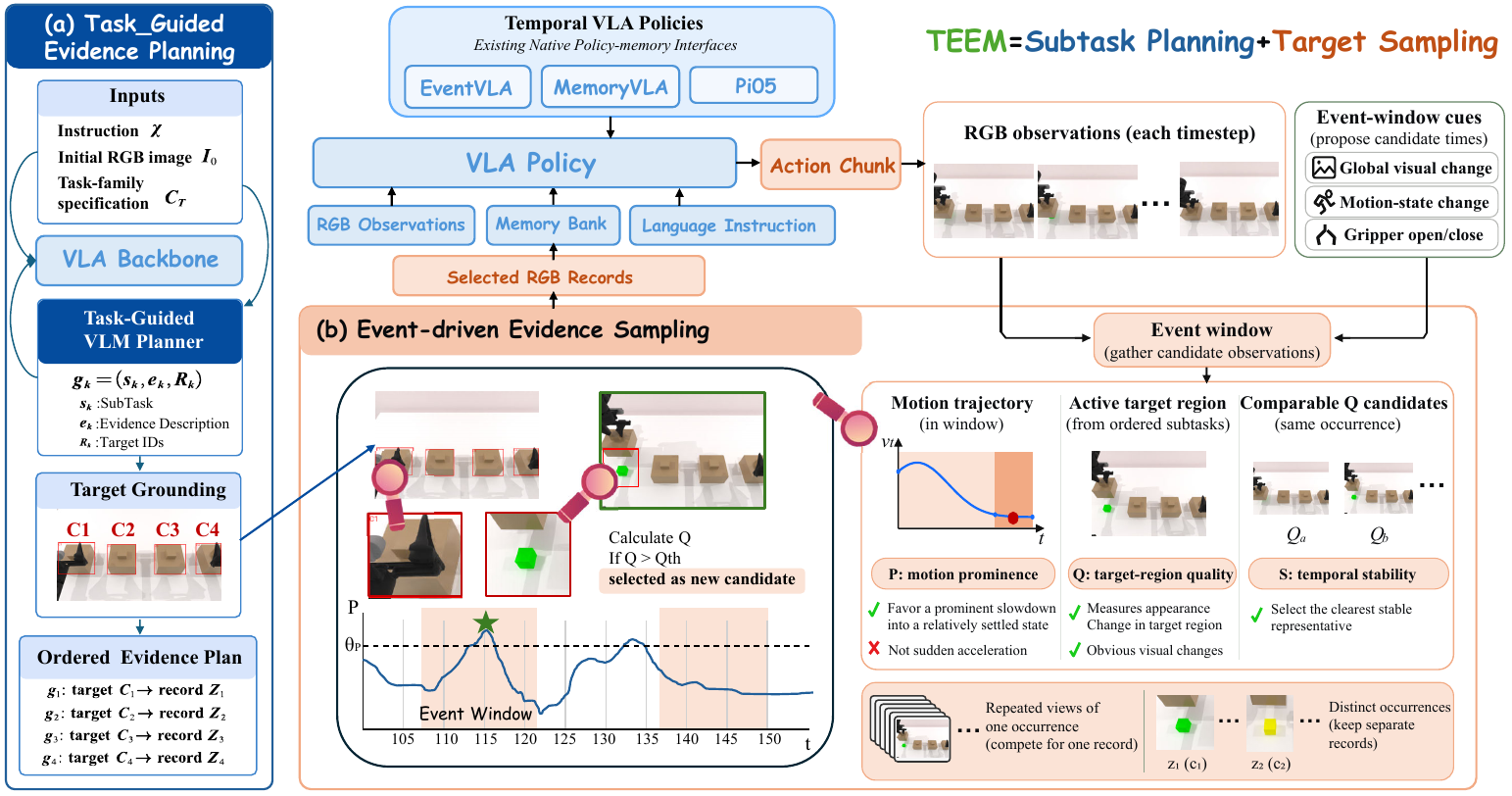}
  \caption{\textbf{Overview of the TEEM framework.} 
  (a) At initialization, the Planning Module maps the instruction, initial RGB image, and task-family specification to an ordered plan of grounded, subtask-linked targets.
  the (b) Sampling Module uses online observations and calculates event cues to open an event window, then selects one RGB record per plan step using motion prominence $P$, target-region appearance change $Q$, and temporal stability $S$. 
  Selected records enter multiple memory-capable VLA policies through their respective memory interfaces, illustrated for EventVLA, MemoryVLA, and $\pi_{0.5}$.}
  \label{fig:teem-method}
\end{figure*}

\noindent\textbf{Overview.}
Research on hierarchical behavior suggests that complex activities can be organized into smaller interactions, while task-dependent attention directs perception toward objects or regions relevant to the current interaction~\cite{farooqui2012hierarchical,rothkopf2007task}. 
Event-perception research further highlights the role of goals in interpreting perceptual changes~\cite{zacks2007event}.

In TEEM, this attentional focus is represented by a \emph{target}, a region associated with a subtask.
We also realize this cognitive organization through task-guided planning (Sec.~\ref{sec:planning}) and event-driven sampling (Sec.~\ref{sec:events}). (Fig.~\ref{fig:teem-method}).
Planning establishes an ordered set of subtask-linked targets and evidence occurrences. 
Sampling then uses physical event cues to gather candidates and target-conditioned measurements to retain one representative per occurrence, after which the selected RGB records enter a compatible policy-native memory pathway.
This separates initialization-time task interpretation from online evidence acquisition without requiring a semantic VLM call for every candidate observation.

\noindent\textbf{Formulation.}
Given instruction $x$, current observation $o_t$, and historical memory $\mathcal{M}_t$, a memory-capable VLA predicts an action chunk
\begin{equation}
    a_{t:t+H-1}=\pi_\theta(x,o_t,\mathcal{M}_t).
    \label{eq:policy}
\end{equation}
TEEM determines which historical RGB observations populate $\mathcal{M}_t$;
the receiving policy retains its own visual encoding, memory reader, and action-generation mechanism. 
We refer to the use of subtask-linked targets to organize this acquisition process as \emph{task-guided evidence focusing}.

\subsection{Task-Guided Evidence Planning}
\label{sec:planning}

At initialization, TEEM uses a VLM planner to organize the task into an ordered set of evidence-bearing subtasks.
A subtask specifies an interaction to be performed, while its target identifies the object or region whose visual state may provide useful evidence. 
A target may be associated with multiple subtasks; each subtask receives its own evidence slot so that distinct evidence requirements remain separate and ordered.

The planner uses two initialization-time passes with distinct roles. 
In the first pass, the VLM interprets the instruction and initial RGB image to decompose the task into ordered subtasks, assign target IDs to relevant objects or regions, and describe the visual evidence to retain for each subtask. 
In the second pass, the planner refines this proposal using the human-authored task-family specification $\mathcal{C}_\tau$.
The specification regularizes the plan structure, target-ID roles, and configured slots or parameterized sequences, while the VLM provides the episode-specific task and scene interpretation. The resulting plan is
\begin{equation}
\begin{aligned}
& \mathcal{G} = \operatorname{Plan}(\mathcal{C}_\tau, x, I_0) = (g_1,\ldots,g_N), \\
& g_k = (s_k, e_k, \mathcal{R}_k), \quad k = 1,\ldots,N,
\end{aligned}
\label{eq:plan}
\end{equation}
where $s_k$ identifies a subtask, $e_k$ describes the visual evidence to retain, and $\mathcal{R}_k$ contains the target IDs associated with that subtask. 
Each evidence-bearing plan node is assigned one evidence slot $z_k$. 
Thus, when the same target appears in multiple subtasks, each subtask retains a distinct slot in the prescribed order.

GroundingDINO maps the target IDs in the initial image $I_0$ to image regions $\{\Omega_r\}$. 
These grounded regions are associated with the corresponding plan nodes and used during online sampling. 
The planner therefore establishes which target regions to monitor at initialization, while subsequent interaction determines when informative observations become available.

\paragraph{Advancing the evidence focus.}
The ordered plan also provides a lightweight progress rule. 
Let $t_k^{\mathrm{conf}}$ be the first step at which $z_k$ contains an admissible representative, and let $\hat r_n$ denote the target ID associated with the $n$-th event window. 
We track the running-best target quality and its latest improvement:
\begin{equation}
\begin{aligned}
& \Gamma_k(t) = \max_{u\in[t_k^{\mathrm{conf}},t]} Q_u^{(k)}, \\
& t_k^{\mathrm{best}}(t_k^{\mathrm{conf}}) = t_k^{\mathrm{conf}}, \\
& t_k^{\mathrm{best}}(t) =
\begin{cases}
t, & \Gamma_k(t) - \Gamma_k(t-1) > \epsilon_Q, \\
t_k^{\mathrm{best}}(t-1), & \text{otherwise.}
\end{cases}
\end{aligned}
\label{eq:quality-trend}
\end{equation}
The focus advances when a new window belongs to another target or when the current target has produced no new best quality for $B=100$ steps:
\begin{equation}
    T_n^{(k)}=H_n^{(k)}\,\mathbb{1}\!\left[
    \hat r_n\notin\mathcal{R}_k\;\lor\;
    t_n-t_k^{\mathrm{best}}(t_n)\geq B\right],
    \label{eq:switch}
\end{equation}
where $H_n^{(k)}$ indicates that $z_k$ is nonempty and $t_n$ is the availability time of the event-window update. 
When $T_n^{(k)}=1$, the cursor locks the current evidence slot and activates the next plan node. The ordered target assignments therefore support lightweight online subtask switching without repeatedly invoking the VLM to determine the active subtask.

\subsection{Event-Driven Evidence Sampling}
\label{sec:events}

Once the planner has assigned a target to the active subtask, semantic interpretation has established the direction of attention, but it has not determined which observation should be stored. 
The robot must still acquire evidence while the interaction unfolds. 
From the target perspective, we formulate two properties for evidence sampling. 
\emph{Target importance} makes candidate evaluation depend on the target regions associated with the active subtask, so unrelated scene changes do not directly determine which record is retained. 
\emph{Target sparsity} assigns one evidence slot to each plan node and lets candidate observations for that node compete to update the same slot. 
Guided by these properties, TEEM first uses physical--visual event cues to propose candidate windows, then applies target-conditioned measurements to select the RGB observation to retain.

\paragraph{Target importance.}
A global visual change, a transition in the motion state, or a gripper opening or closing marks a possible interaction boundary. 
These cues open a window around the interaction, following event-perception accounts in which continuous activity is organized around perceptual changes~\cite{zacks2007event} and robotic memory systems use kinematic or visual signals to propose candidate moments~\cite{zeng2026kemo,yang2026eventvla}. 
These signals provide a low-cost temporal trigger rather than a semantic judgment about task relevance. 
We therefore define an event window that collects observations around a possible interaction boundary. With proprioceptive motion signal $m_t$, global visual-change signal $v_t$, and gripper trigger $h_t$, the window opens when
\begin{equation}
    E_t=\mathbb{1}\!\left[h_t=1\;\lor\;
    m_t>\theta_m^{\mathrm{on}}\;\lor\;
    v_t>\theta_v^{\mathrm{on}}\right]=1.
    \label{eq:window-open}
\end{equation}
The window only determines when candidate observations become available for evaluation. Whether those observations contain useful task evidence is decided later by the active target and its grounded target regions.

Once a window is open, TEEM does not retain every visually changed frame. 
It analyzes the local motion trajectory and favors an observation near the transition from active movement to a relatively settled interaction state. 
In practice, this preference avoids selecting a highly transient frame at the peak of motion and instead captures the state that remains after the interaction has produced a visible change. 
For normalized proprioceptive velocity $\nu_t\in[0,1]$ and history length $W_m$, we use
\begin{equation}
    m_t=(1-\nu_t)\max_{j\in[t-W_m+1,t]}[\nu_{j-1}-\nu_j]_+,
    \label{eq:motion-signal}
\end{equation}
and retain a local candidate $c$ when its prominence over the local motion baseline exceeds $\theta_P$:
\begin{equation}
    P_c=\frac{[m_c-\mu_c]_+}{1.4826\,\operatorname{MAD}_c+\varepsilon},\qquad
    A_c=\mathbb{1}[P_c\geq\theta_P].
    \label{eq:motion-prominence}
\end{equation}
Motion prominence alone, however, does not indicate whether the active target has become visually informative. 
We therefore evaluate each eligible RGB observation within the grounded target regions.
For region $\Omega_r$, the target-conditioned appearance score is
\begin{equation}
\begin{aligned}
& Q_{c,r} = \frac{1}{|\Omega_r|}\sum_{p\in\Omega_r}
\phi\!\left(I_0(p), I_c(p)\right), \\
& Q_c^{(k)} = \mathcal{A}_k\!\left(\{Q_{c,r} : r \in \mathcal{R}_k\}\right).
\end{aligned}
\label{eq:target-quality}
\end{equation}
where $\phi$ is an illumination-normalized local appearance change and $\mathcal{A}_k$ follows the target specification for single or composite regions. 
The target-conditioned score compares the reference and candidate appearances after reducing sensitivity to global illumination changes. In our implementation, the pixelwise comparison can use an optional robot-occlusion mask to exclude pixels covered by the arm or gripper, while normalization by the full region area prevents a small visible fragment from receiving a high score simply because the remaining region is masked. The regional scores are then aggregated according to whether the plan specifies a single target or a composite target.

\paragraph{Target sparsity.}
Evidence for an extended interaction is not naturally tied to one arbitrary instant. Several observations may depict the same event as the robot approaches, reveals, or settles on the target. 
Retaining all of these observations would spend the bounded memory on repeated views, whereas committing the first admissible observation could preserve a transient or partially occluded state. 
TEEM models this trade-off explicitly by letting candidates assigned to the same planned occurrence compete for one record.
It prefers the one that remains stable across the following observations:
\begin{equation}
    S_{c,r}=\frac{1}{J|\Omega_r|}\sum_{j=1}^{J}\sum_{p\in\Omega_r}
    M_c(p)\mathbb{1}\!\left[\|I_{c+j}(p)-I_c(p)\|_2<\theta_s\right],
    \label{eq:stability}
\end{equation}
and aggregates the regional values using the same target rule as $Q$.
Let $\mathcal{B}_k(t)$ contain candidates that pass $A_c$, satisfy $Q_c^{(k)}\geq\theta_{Q,k}$, and have all required confirmation observations available. 
We select
\begin{equation}
\begin{aligned}
& Q_k^{\max}(t) = \max_{c\in\mathcal{B}_k(t)} Q_c^{(k)}, \\
& c_k^*(t) = \arg\max_{c:\,Q_c^{(k)}\geq Q_k^{\max}(t)-\epsilon_Q} S_c^{(k)}.
\end{aligned}
\label{eq:representative}
\end{equation}
Quality has priority, while stability resolves near-ties. 
A later candidate can replace the current representative in $z_k$ while plan node $k$ remains active.
Once the cursor advances, the slot is locked; subsequent plan nodes retain separate slots even when they use the same target ID.

\subsection{Policy Memory Interface}
\label{sec:policy-interface}

TEEM is an external evidence-selection module that determines which observations enter memory while leaving visual encoding, memory reading, and action generation to the receiving VLA.
Its output is an ordered set of selected full-RGB observations together with source-time and occurrence metadata.
Plan metadata, target IDs, and revision states remain on the selector side; the receiving policy only needs an existing historical-image or memory input through which it can encode and read the selected records. 
This boundary lets the same acquisition logic be evaluated with different memory-capable policies without replacing their action heads.

For an event-indexed visual-history policy such as EventVLA, the selected records occupy its visual history slots in place of the policy's native selection decisions~\cite{yang2026eventvla}.
For a feature-memory policy such as MemoryVLA, each selected RGB image is encoded using the policy's native visual encoders. 
The resulting features are written to dedicated, protected TEEM evidence slots, separate from ordinary native entries~\cite{shi2025memoryvla}.
For each memory bank with total capacity $K$, we reserve $E$ entries for TEEM:
\begin{equation}
|\mathcal{M}^{\mathrm{native}}_t| \leq K - E,
\quad
|\mathcal{M}^{\mathrm{TEEM}}_t| \leq E,
\quad
|\mathcal{M}_t| \leq K.
\label{eq:memory-budget}
\end{equation}
The policy reads the native and TEEM entries through its memory pathway, while the reserved TEEM entries are excluded from ordinary native-memory compression or replacement.
The selector is shared, but the receiving policy retains its own representation and reader; compatibility therefore means using an existing memory pathway rather than claiming zero-adaptation support for every VLA.

\section{Experiments}
\label{sec:experiments}
In this section, we evaluate TEEM as a task-guided evidence-sampling module for long-horizon manipulation through four questions:
\textbf{Q1}: Can TEEM capture task-relevant keyframes and improve long-horizon task success?
\textbf{Q2}: Can the same sampling module work with different policy-native memory pathways?
\textbf{Q3}: How do target importance, target sparsity, and their associated sampling criteria affect keyframe selection?
\textbf{Q4}: Can TEEM be applied to long-horizon manipulation on a physical robot?

\begin{figure*}[t]
  \centering
  \includegraphics[width=\textwidth]{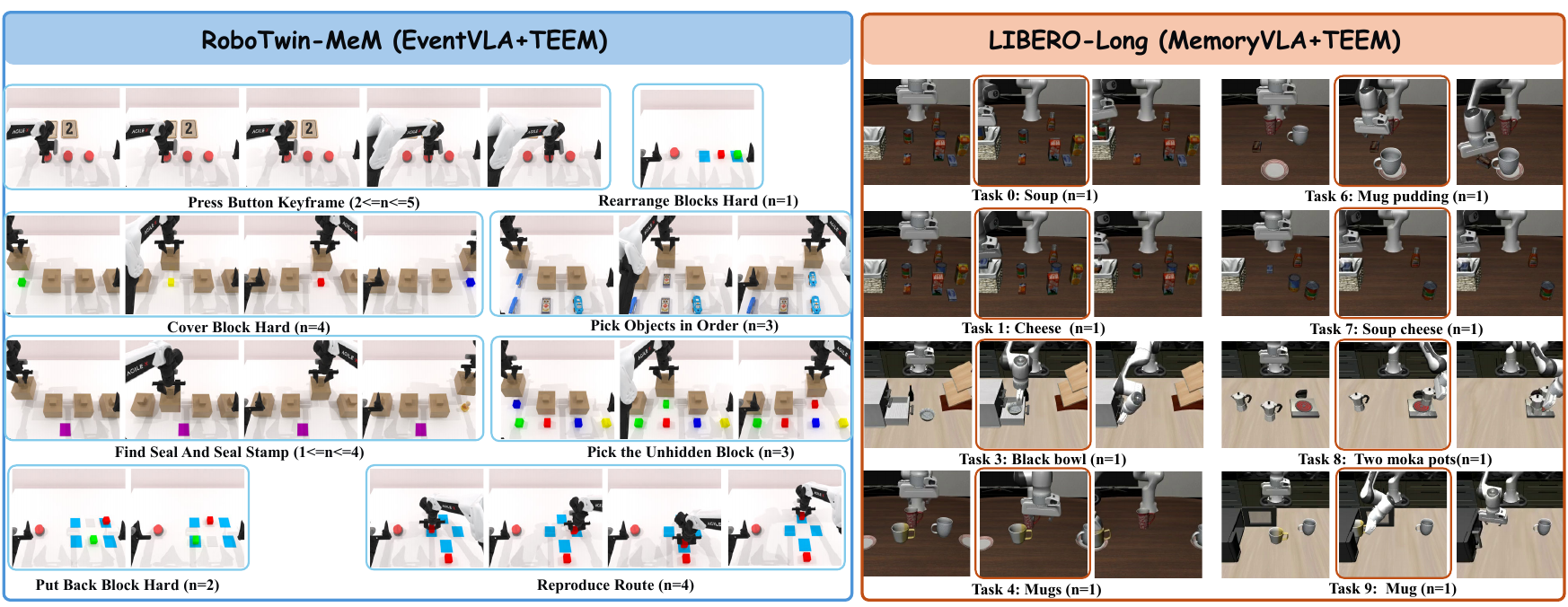}
  \caption{\textbf{Representative tasks from RoboTwin-MeM and LIBERO-Long.}
  In RoboTwin-MeM, we only show task-linked keyframes; $n$ denotes the number of required keyframes. 
  In LIBERO-Long, each example shows the initial observation, the keyframe, and the final task state. }
  \label{fig:benchmark-overview}
\end{figure*}

The Main Results address Q1 through task success and keyframe-level analyses, and Q2 by integrating TEEM with EventVLA and MemoryVLA. The Ablation Studies examine the sampling mechanisms associated with Q3. The Real-World Robot Evaluation applies TEEM to physical-robot tasks to address Q4.

\subsection{Experimental Setup}
\label{sec:exp-setup}

\paragraph{Benchmarks and policies.}
We evaluate two benchmark-policy settings with distinct temporal-memory pathways.
RoboTwin-MeM contains eight tasks requiring intermediate visual evidence and is paired with EventVLA, which stores event-indexed RGB observations in its native memory~\cite{yang2026eventvla}.
LIBERO-Long provides Long-10 and Long-90 suites. 
We evaluate MemoryVLA on Long-10 and on a selected 10-task subset of Long-90; its native pathway encodes and retrieves past observations through feature memory~\cite{shi2025memoryvla}.
For LIBERO, both the RGB observations used for keyframe sampling and the initial image used for target-region grounding come from the fixed oblique third-person \texttt{agentview} camera.
RoboTwin-MeM uses a fixed overhead \texttt{cam\_high} view for its temporal-memory pathway. 
Figure~\ref{fig:benchmark-overview} shows representative examples; our Long-90 results cover only the selected subset, not the full suite.

\paragraph{Implementation Details.}
TEEM does not update the VLA weights or require policy retraining. 
For each paired native-versus-TEEM evaluation, we keep the policy checkpoint, action-generation settings, and evaluation conditions fixed.
The MemoryVLA comparison also uses a matched total memory budget. 
Detailed policy, task, and integration configurations are provided in the supplementary material.
TEEM parameters are specified in task-family configurations: applicable thresholds may be calibrated per task and are then held fixed during evaluation. 
Thus, our claim is training-free integration with two existing memory pathways, not universal zero-shot compatibility with arbitrary VLA policies.

\begin{wrapstuff}[type=table,width=0.4\textwidth,r]
\centering
\footnotesize
\caption{Paired MemoryVLA results on LIBERO-Long (SR \%, 50 initial states).
Suite headings report unweighted mean SR as Native (MemoryVLA)\,/\,TEEM (MemoryVLA + TEEM);
bold entries match or exceed Native.}
\label{tab:libero-long-results}
\vspace*{5pt}
\scriptsize
\setlength{\tabcolsep}{2.5pt}
\renewcommand{\arraystretch}{1.05}
\begin{tabular}{l*{5}{c}}
\toprule
\multicolumn{6}{l}{\emph{LIBERO-Long-10} \hfill Avg.\ SR: 90.0\% / \textbf{91.8\%}} \\
\cmidrule(lr){1-6}
\rowcolor{methodRow}Method & Soup & Cheese & \shortstack{Moka\\pot} & \shortstack{Black\\bowl} & Mugs \\
\midrule
Native & 90\% & 100\% & 100\% & 98\% & 96\% \\
TEEM   & \textbf{90\%} & \textbf{100\%} & \textbf{100\%} & \textbf{98\%} & \textbf{98\%} \\
\addlinespace[2pt]
\rowcolor{methodRow}Method & Book & \shortstack{Mug\\pudding} & \shortstack{Soup\\cheese} & \shortstack{Two moka\\pots} & Mug \\
\midrule
Native & 90\% & 90\% & 94\% & 62\% & 80\% \\
TEEM   & \textbf{90\%} & \textbf{92\%} & \textbf{94\%} & \textbf{70\%} & \textbf{86\%} \\
\midrule
\multicolumn{6}{l}{\emph{LIBERO-Long-90 (part)} \hfill Avg.\ SR: 93.4\% / \textbf{94.2\%}} \\
\cmidrule(lr){1-6}
\rowcolor{methodRow}Method & \shortstack{Bowl\\atop} & \shortstack{Rear\\butter} & \shortstack{Front\\butter} & Pudding & Bowl \\
\midrule
Native & 100\% & 100\% & 96\% & 100\% & 100\% \\
TEEM   & \textbf{100\%} & \textbf{100\%} & \textbf{96\%} & \textbf{100\%} & \textbf{100\%} \\
\addlinespace[2pt]
\rowcolor{methodRow}Method & \shortstack{Pan\\(K.\,3)} & \shortstack{Open/\\close} & \shortstack{Pan\\(K.\,9)} & \shortstack{Stack\\L$\rightarrow$R} & \shortstack{Stack\\R$\rightarrow$L} \\
\midrule
Native & 72\% & 96\% & 96\% & 98\% & 76\% \\
TEEM   & \textbf{78\%} & \textbf{96\%} & \textbf{98\%} & \textbf{98\%} & \textbf{76\%} \\
\bottomrule
\end{tabular}
\vspace{-12pt}
\end{wrapstuff}

\subsection{Main Results}
\label{sec:main-results}

\noindent\textbf{Evaluation on LIBERO-Long.}
We first evaluate MemoryVLA on LIBERO-Long, a collection of tabletop spatial-manipulation tasks. 
LIBERO-Long tasks expose ordered subtasks linked to multiple target regions, requiring joint evidence from multiple target regions. 
On both evaluated LIBERO-Long suites, MemoryVLA + TEEM matches or improves the native success rate on every task.
On Long-10, TEEM raises the success rate for \emph{Two Moka Pots} from 62\% to 70\%; 
the task requires placing two pots at the stove target. 
It also improves \emph{Mug} from 80\% to 86\%, where successful completion requires placing the mug and then closing the microwave. 
These gains are consistent with retaining target-region evidence as the active subtask changes. 
On the selected Long-90 subset, the average success rate rises from 93.4\% to 94.2\%, including a 6-point gain on \emph{Pan on Stove} (Kitchen 3), from 72\% to 78\%.

We further analyzed the timing of retained evidence.
LIBERO-Long rollouts average 214.6 environment steps. 
These relatively short histories give MemoryVLA's native compression mechanism a manageable sequence to summarize, helping establish a strong baseline and limiting the average headroom for TEEM. 
The benefit is clearer when important evidence appears early: for \emph{Pan (K.3)}, the median selected evidence timestep is 93 in a median 250-step trajectory, and TEEM improves success from 72\% to 78\%. 
This result illustrates the value of preserving an early target-specific observation that can inform decisions throughout the remaining trajectory.

\begin{table*}[t]
  \centering
  \scriptsize
  \setlength{\tabcolsep}{3pt}
  \renewcommand{\arraystretch}{1.0}
  \definecolor{teemAblation}{RGB}{150,150,150}
  \definecolor{groupRow}{RGB}{214,226,245}
  \definecolor{ablationRow}{RGB}{255,241,226}
  \caption{RoboTwin-MeM success rates (\%) with EventVLA, ordered by the required number of keyframes $n$. 
  Previously reported baselines are taken from EventVLA~\cite{yang2026eventvla}; 
  the Official EventVLA, EventVLA + TEEM, and ablation rows report our evaluations, using same checkpoint from EventVLA~\cite{yang2026eventvla}. 
  Avg. is the macro-average over the eight tasks.
  TEEM ablations remove the target-region quality $Q$, motion-prominence $P$, and temporal-stability $S$ scores, respectively (shown in gray). 
  Bold values mark the best result in each task.}
  \label{tab:robotwin-mem-results}
  \begin{tabular}{l*{9}{c}}
    \toprule
    Method & \multicolumn{1}{c}{$n=1$} & \multicolumn{1}{c}{$n=2$} &
    \multicolumn{2}{c}{$n=3$} & \multicolumn{3}{c}{$n=4$} &
    \multicolumn{1}{c}{$n=5$} & Avg. \\
    \cmidrule(lr){2-2}\cmidrule(lr){3-3}\cmidrule(lr){4-5}
    \cmidrule(lr){6-8}\cmidrule(lr){9-9}\cmidrule(lr){10-10}
    & \shortstack{Rearrange\\Blocks Hard}
    & \shortstack{Put Back\\Block Hard}
    & \shortstack{Pick Objects\\in Order}
    & \shortstack{Pick the\\Unhidden Block}
    & \shortstack{Cover\\Blocks Hard}
    & \shortstack{Find Seal\\Stamp}
    & \shortstack{Reproduce\\Route}
    & \shortstack{Press Button\\Keyframe}
    & \\
    \midrule
    \rowcolor{groupRow}\multicolumn{10}{l}{\emph{Non-memory-based vision-language-action models}} \\
    $\pi_{0.5}$ & 20\% & 19\% & 1\% & 14\% & 0\% & 8\% & 0\% & 0\% & 7.8\% \\
    QwenOFT & 3\% & 26\% & 0\% & 0\% & 0\% & 0\% & 0\% & 1\% & 3.8\% \\
    \addlinespace[1pt]
    \rowcolor{groupRow}\multicolumn{10}{l}{\emph{Dual-system memory-based models}} \\
    MemER & 32\% & 4\% & 12\% & 2\% & 0\% & 26\% & 3\% & 5\% & 10.5\% \\
    Mem-0 & 0\% & 0\% & 0\% & 0\% & 0\% & 0\% & 0\% & 0\% & 0.0\% \\
    \addlinespace[1pt]
    \rowcolor{groupRow}\multicolumn{10}{l}{\emph{End-to-end memory-based vision-language-action models}} \\
    MemoryVLA (OpenVLA) & 12\% & 0\% & 0\% & 1\% & 0\% & 10\% & 2\% & 14\% & 4.9\% \\
    MemoryVLA (QwenOFT) & 39\% & 0\% & 1\% & 9\% & 1\% & 11\% & 0\% & 25\% & 10.8\% \\
    EventVLA  & 72\% & 100\% & 40\% & 86\% & 89\% & 74\% & 84\% & 72\% & 77.1\% \\
    EventVLA + TEEM & \textbf{80\%} & \textbf{100\%} & \textbf{46\%} & \textbf{86\%} & \textbf{100\%} & \textbf{76\%} & \textbf{100\%} & \textbf{100\%} & \textbf{86.0\%} \\
    \midrule
    \rowcolor{ablationRow}\multicolumn{10}{l}{\emph{TEEM ablations}} \\
    \textcolor{teemAblation}{w/o Target ($Q$)} & \textcolor{teemAblation}{0\%} & \textcolor{teemAblation}{0\%} & \textcolor{teemAblation}{0\%} & \textcolor{teemAblation}{0\%} & \textcolor{teemAblation}{0\%} & \textcolor{teemAblation}{0\%} & \textcolor{teemAblation}{0\%} & \textcolor{teemAblation}{90\%} & \textcolor{teemAblation}{11.3\%} \\
    \textcolor{teemAblation}{w/o Motion ($P$)} & \textcolor{teemAblation}{6\%} & \textcolor{teemAblation}{0\%} & \textcolor{teemAblation}{24\%} & \textcolor{teemAblation}{4\%} & \textcolor{teemAblation}{18\%} & \textcolor{teemAblation}{34\%} & \textcolor{teemAblation}{0\%} & \textcolor{teemAblation}{0\%} & \textcolor{teemAblation}{10.8\%} \\
    \textcolor{teemAblation}{w/o Stability ($S$)} & \textcolor{teemAblation}{22\%} & \textcolor{teemAblation}{20\%} & \textcolor{teemAblation}{46\%} & \textcolor{teemAblation}{86\%} & \textcolor{teemAblation}{100\%} & \textcolor{teemAblation}{76\%} & \textcolor{teemAblation}{16\%} & \textcolor{teemAblation}{100\%} & \textcolor{teemAblation}{58.3\%} \\
    \bottomrule
  \end{tabular}
\end{table*}

\noindent\textbf{Evaluation on RoboTwin-MeM.}
We next evaluate EventVLA on RoboTwin-MeM, whose longer-horizon tasks require task-dependent numbers of intermediate visual records, ranging from one to five. 
As shown in Table~\ref{tab:robotwin-mem-results}, replacing EventVLA's native KEM selection with TEEM improves the average success rate from 77.1\% to 86.0\% without additional training, achieving the best average result among the reported methods on RoboTwin-MeM.
Notably, it reaches 100\% success on several tasks, including \texttt{reproduce\_route} and \texttt{cover\_blocks\_hard}. 
These tasks have clearly defined subtasks and explicit target regions. 
By combining physical interaction cues with target-region visual changes, TEEM identifies the required keyframes and uses a simple subtask-switching rule to preserve the corresponding evidence in memory.
However, failures remain on tasks such as \texttt{pick\_objects\_in\_order}.
A per-task review of selected runs confirmed that TEEM captured the required keyframe types across all eight task families. 
In some unsuccessful episodes, however, EventVLA repeatedly attempted grasps even after the relevant frame had been written to memory. 
This suggests that these failures arise from downstream policy execution rather than from TEEM's failure to acquire the required visual evidence.
Together, the results show that capturing task-relevant keyframes at the right time makes earlier visual information available to later subtasks and improves long-horizon task success.

\noindent\textbf{The Effect of TEEM.}
Across both long-horizon benchmarks, TEEM improves aggregate task success without additional training. 
On LIBERO-Long, it raises the mean success rate from 90.0\% to 91.8\% on Long-10 and from 93.4\% to 94.2\% on the selected LIBERO-Long-90 subset. On RoboTwin-MeM, it raises the average from 77.1\% to 86.0\%.
TEEM is particularly effective on tasks with clear subtask structure and explicit target regions. 
In these tasks, keyframe information often appears early in the interaction but is needed again in later stages.
Event cues propose candidate time windows, target-region visual quality identifies which observations contain useful information, and the subtask-switching rule preserves the selected records in order. 
The RoboTwin-MeM results also show substantially lower success for the listed MemoryVLA variants, which compress historical observations into memory representations. 
This contrast underscores the value of TEEM — accurately identifying and retaining task-relevant keyframes in long-horizon manipulation.

\noindent\textbf{Generality Across VLA Architectures.}
We assess TEEM's generality by integrating the same external selector with EventVLA and MemoryVLA, which use distinct temporal-memory pathways. 
TEEM retains selected RGB observations as ordered evidence records and integrates them with each policy through its memory interface.
For EventVLA, TEEM directly replaces the native KEM selection pathway. 
For MemoryVLA, we add dedicated TEEM slots to the memory bank. The selected records then follow the existing downstream memory pathway without undergoing the native compression and replacement operations. 
Across both architectures, TEEM improves aggregate success, including on tasks where the native policy has a relatively low success rate.

\subsection{Ablation Studies}
\label{sec:ablations}

We evaluate target importance by replacing target-conditioned quality $Q$ with a whole-image score (\emph{w/o Target}). 
To examine motion-based timing, \emph{w/o $P$} removes prominence-based frame selection and instead selects a candidate after a fixed interval from event-window opening. We assess the representative-selection rule associated with target sparsity by removing temporal-stability ranking $S$.

Removing $Q$ reduces success to zero on seven tasks, while \emph{Press Button Keyframe} retains 90\% success, compared with 100\% for full TEEM. 
This task repeatedly interacts with a single visually salient target, allowing whole-image changes to serve as a useful proxy for frame selection.
The other tasks involve multiple objects or target regions, where global visual changes do not reliably identify which task-relevant state should be retained.

Removing $P$ has a different effect: success on \emph{Press Button Keyframe} falls from 100\% to 0\%. 
Target-conditioned quality indicates what to observe, but does not identify the brief interaction moment; motion prominence provides this timing cue. 
In contrast, \emph{Find Seal Stamp} retains 34\% success without $P$, showing that event-window cues and target-conditioned quality can still identify useful observations in some episodes.

We next examine target sparsity by removing $S$ while retaining one evidence slot per planned subtask. The average success rate falls from 86.0\% to 58.3\%, with marked drops on \emph{Rearrange Blocks Hard} (80\% to 22\%), \emph{Put Back Block Hard} (100\% to 20\%), and \emph{Reproduce Route} (100\% to 16\%). 
The other five tasks retain their full-TEEM success rates. These results show that stability-based ranking is particularly useful when later actions depend on retaining a representative view of a spatial relation or route state, rather than a transient view during interaction.

\subsection{Real-World Robot Evaluation}

\begin{wrapfigure}{r}{0.5\textwidth}
\vspace{-14pt}
\centering
\includegraphics[width=\linewidth]{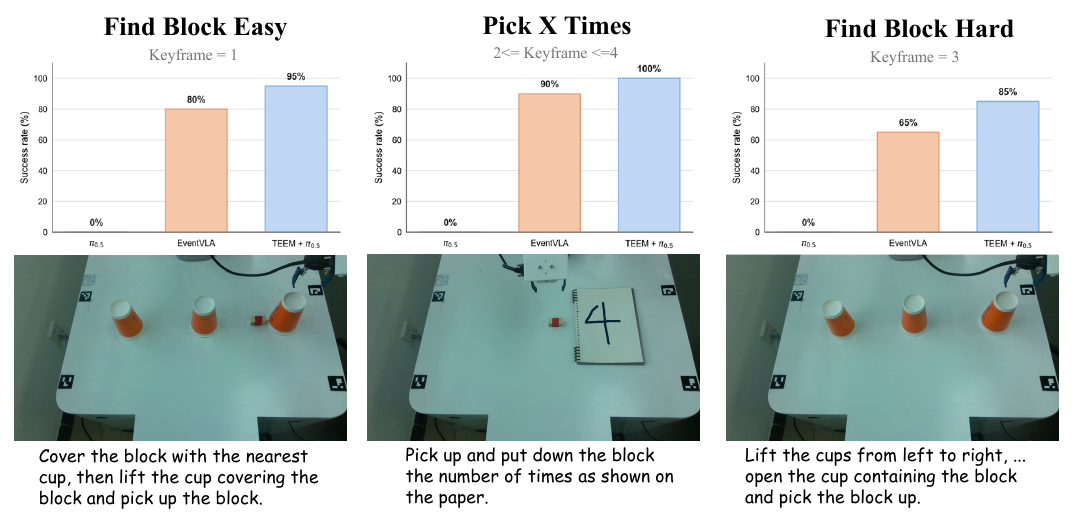}
\caption{\textbf{Real-world evaluation on three long-horizon manipulation tasks.}}
\label{fig:real-world-results}
\vspace{-10pt}
\end{wrapfigure}

We evaluate TEEM on a physical DOBOT CR5V robot using three non-Markovian manipulation tasks: \texttt{Find\_Block\_Easy}, \texttt{Pick\_X\_Time}, and \texttt{Find\_Block\_Hard}. 
The tasks require one, between two and four, and three keyframes, respectively. 
We compare three policies: the memory-free $\pi_{0.5}$ baseline, EventVLA fine-tuned with a $\pi_{0.5}$ backbone, and $\pi_{0.5}$ augmented with TEEM.

As shown in Fig.~\ref{fig:real-world-results}, the memory-free $\pi_{0.5}$ baseline achieves 0\% success on all three tasks. 
Fine-tuned EventVLA achieves success rates of 80\%, 90\%, and 65\%, respectively, for a mean of 78.3\% across the tasks.
With TEEM, $\pi_{0.5}$ reaches 95\%, 100\%, and 85\%, yielding a mean success rate of 93.3\%. 
These results show that TEEM substantially improves physical-robot performance on the evaluated long-horizon tasks. 
By combining trajectory-based physical cues with target-region visual assessment, TEEM selects and retains task-relevant keyframes that help the policy use information from earlier interactions across ordered subtasks.

\section{Conclusion}
We introduced TEEM, a task-guided evidence-sampling module for long-horizon manipulation. 
TEEM uses an initialization-time plan to associate subtasks with target regions and evidence requirements, then samples visual observations online using physical event cues, motion prominence, target-conditioned visual quality, and temporal stability. 
Its policy-external design allows selected RGB records to enter different VLA memory pathways without replacing the receiving policy's action-generation mechanism.

Across LIBERO-Long, RoboTwin-MeM, and three physical-robot tasks, TEEM improves or preserves task success while supporting both feature-memory and event-indexed visual-history pathways. 
The LIBERO results show consistent per-task non-degradation, while the RoboTwin-MeM results and ablations highlight the roles of target-conditioned scoring and motion-based candidate timing. 
Real-robot results further show that TEEM can support long-horizon execution on a DOBOT CR5V. 
The limitations and future directions are discussed in the supplementary material.

\bibliographystyle{unsrt}  
\bibliography{references}  

\clearpage
\setcounter{page}{1}

\section{Implementation Details}
\label{sec:supp-implementation}

\subsection{Task-Guided Evidence Planning}
\label{sec:supp-task-planning}

TEEM uses a shared prompt template to interpret the task instruction and initial RGB observation. 
The prompt first asks the VLM to identify candidate target regions and produce a coarse ordered task plan. 
Its output contains \texttt{available\_roi\_ids}, which lists the region IDs, labels, and grounding queries, and \texttt{steps}, which gives a high-level summary and evidence description for each step.
This common schema provides a consistent starting point across task families. 
We then manually refine the plan where necessary to define task-appropriate subtask boundaries and target assignments.

\paragraph{LIBERO-Long.}
Typically, for LIBERO-Long, the VLM-generated plan can be used directly because the task descriptions are explicit and the required interactions are relatively concise. 
For example, for task 0, the VLM identifies the basket, the alphabet soup can, and the tomato sauce bottle as candidate regions.
It then generates two ordered subtasks, placing the soup can into the basket followed by placing the tomato sauce bottle into the basket.
The associated evidence conditions describe the completed placement of each object.

\begin{lstlisting}[style=cfgcompact, language=Python,
caption={Basket-packing task configuration.}]
available_roi_ids:
R1: woven basket with silver lining
R2: can of alphabet soup
R3: bottle of tomato sauce
steps:
0: place the alphabet soup can into the basket
Evidence: the soup can is inside the basket
1: place the tomato sauce bottle into the basket
Evidence: the sauce bottle is inside the basket
\end{lstlisting}

This output already provides the required target regions, subtask order, and evidence descriptions, so no additional manual restructuring is applied for this LIBERO-Long example.

\paragraph{RoboTwin-MeM.}
For RoboTwin-MeM, the VLM also produces the correct object order and overall task stages, but its initial plan can be more fine-grained than the keyframe sequence required by the native EventVLA memory pathway. For example, for \texttt{cover\_blocks\_hard}, the VLM identifies four covers from left to right and expands each inspection into three separate steps: opening the cover, inspecting the revealed property, and closing the cover. It then adds a final step for reopening the covers in the recorded red, green, blue, and yellow order. This initial output therefore contains 13 steps.

The task-family guide further organizes these steps into five evidence-aligned nodes. The first four nodes correspond to inspecting and recording the revealed property under covers C1--C4, while the fifth node aggregates the ordered reopening of all four covers.

\begin{lstlisting}[style=cfgcompact, language=Python,
caption={Cover-blocks task configuration.}]
human_planned_steps:
S1: open C1, inspect and record its revealed property, then close C1
S2: open C2, inspect and record its revealed property, then close C2
S3: open C3, inspect and record its revealed property, then close C3
S4: open C4, inspect and record its revealed property, then close C4
S5: reopen C1, C2, C3, C4 in the recorded color order
\end{lstlisting}

The task-family guide also replaces the initial region identifiers with fixed identifiers
\texttt{C1}--\texttt{C4} and uses the common grounding query \texttt{cover}. 
Descriptions such as ``cover with black cone'' are retained as initial visual distinctions proposed by the VLM, while the final region grounding follows the task-family guide. This refinement preserves the VLM-generated ordering while matching the required keyframe sequence and EventVLA's native memory pathway.

This difference reflects the structure of the two benchmarks. 
LIBERO-Long tasks have sufficiently clear descriptions for the VLM-generated plans to be used directly. 
RoboTwin-MeM tasks, in contrast, follow EventVLA's native subtask structure, so we manually refine the plans to align their subtask boundaries and targets with the required keyframe sequence.

\subsection{Event-Driven Evidence Sampling}
\label{sec:supp-sampling}

After task planning, TEEM samples visual evidence online for the active plan node.
Physical and visual event cues first open a candidate window. 
TEEM then applies the task-configured motion gate $P$ and the visual event gate $D$ to identify eligible candidate frames. 
It evaluates the remaining candidates using target-conditioned quality $Q$ over the active node's ROI set.
Then TEEM uses temporal-stability ranking $S$ to select among candidates with comparable $Q$ scores. 
The selected full-frame RGB observation is written to the evidence slot associated with the active plan node.

For all evaluated tasks in LIBERO-Long-10 and the selected LIBERO-Long-90 subset, we use the same sampling thresholds: motion prominence $P \geq 0.40$ and target-region quality $Q \geq 0.12$. 
The ROI set used to compute $Q$ is task-specific and follows the grounded targets associated with the active subtask. 
For example, it includes the soup can and basket for LIBERO-Long-10 task 0, the stove for task 2, and the two bowls for LIBERO-Long-90 tasks 63 and 64.

RoboTwin-MeM uses task-specific thresholds for motion prominence and visual event filtering.
Table~\ref{tab:supp-robotwin-sampling} summarizes these settings.

\begin{wraptable}{r}{0.40\textwidth}
\vspace{-14pt}
\centering
\footnotesize
\caption{Task-specific sampling configuration for RoboTwin-MeM. $P$ is the
motion gate, $D$ is the visual event gate, and $Q$ is the target-region
quality gate.}
\label{tab:supp-robotwin-sampling}
\vspace*{5pt}
\scriptsize
\setlength{\tabcolsep}{4pt}
\renewcommand{\arraystretch}{1.15}
\begin{tabular}{lccc}
\toprule
Task & $P$ & $D$ & $Q$ \\
\midrule
\texttt{cover\_blocks\_hard}           & 0.40 & 0.60 & 0.30 \\
\texttt{press\_button\_keyframe}       & 0.20 & 0.59 & 0.02 \\
\texttt{put\_back\_block\_hard}        & 0.30 & 0.60 & 0.12 \\
\texttt{reproduce\_route}              & 0.40 & 0.10 & 0.12 \\
\texttt{pick\_objects\_in\_order}      & 0.40 & 0.60 & 0.30 \\
\texttt{pick\_the\_unhidden\_block}    & 0.15 & 0.05 & 0.25 \\
\texttt{find\_seal\_and\_seal\_stamp}  & 0.40 & 0.60 & 0.25 \\
\texttt{rearrange\_blocks\_hard}       & 0.40 & 0.05 & 0.30 \\
\addlinespace[1.5pt]
\bottomrule
\end{tabular}
\vspace{-12pt}
\end{wraptable}

\subsection{Policy Memory Interfaces}
\label{sec:supp-interface}

TEEM outputs an ordered set of selected full-RGB observations. 
The receiving VLA policy retains its own visual encoder, memory reader, and action generation mechanism.

For EventVLA, TEEM replaces the native KEM selection decision and writes the selected RGB records into the policy's event-indexed visual-history pathway.
The EventVLA policy is used with the same checkpoint and action-generation settings as the native comparison.

For MemoryVLA, each selected RGB image is processed by the policy's native visual encoder.
The resulting features are written into dedicated TEEM evidence slots in the memory bank. 
These slots remain separate from ordinary native entries and are excluded from native compression or replacement during the corresponding evaluation. 
The total memory budget is kept fixed.

\section{Evaluation Protocols}
\label{sec:supp-protocols}

\subsection{LIBERO-Long}
\label{sec:supp-libero}

We evaluate all ten tasks in LIBERO-Long-10 and ten selected tasks from
LIBERO-90, using 50 initial states per task. Table~\ref{tab:supp-libero-tasks}
lists the scene and instruction for each evaluated task. The fixed oblique
\texttt{agentview} camera provides the initial image for target grounding
and the RGB observations for keyframe sampling.

\begin{table}[t]
\centering
\small
\setlength{\tabcolsep}{5pt}
\definecolor{groupRow}{RGB}{214,226,245}
\definecolor{zebraRow}{RGB}{242,244,248}
\caption{Evaluated LIBERO tasks, their shorthand names, and mean steps per
episode. IDs are zero-based within each suite. All episodes are over 50
initial states.}
\label{tab:supp-libero-tasks}
\begin{tabularx}{\linewidth}{@{}cc@{\hspace{8pt}}c@{\hspace{10pt}}>{\raggedright\arraybackslash}X@{}}
\toprule
ID & Name & \shortstack{Mean\\steps} & Task Description \\
\midrule
\rowcolor{groupRow}\multicolumn{4}{@{}l}{\emph{LIBERO-Long-10}} \\
0 & Soup & 295 & Put the alphabet soup and tomato sauce in the basket. \\
\rowcolor{zebraRow}1 & Cheese & 243 & Put the cream cheese box and butter in the basket. \\
2 & Moka pot & 234 & Turn on the stove and place the moka pot on it. \\
\rowcolor{zebraRow}3 & Black bowl & 227 & Put the black bowl in the bottom drawer and close it. \\
4 & Mugs & 230 & Put the white mug on the left plate and the yellow-and-white mug on the right plate. \\
\rowcolor{zebraRow}5 & Book & 210 & Put the book in the rear compartment of the caddy. \\
6 & Mug pudding & 253 & Put the white mug on the plate and the pudding to its right. \\
\rowcolor{zebraRow}7 & Soup cheese & 264 & Put the alphabet soup and cream cheese box in the basket. \\
8 & Two moka pots & 451 & Put both moka pots on the stove. \\
\rowcolor{zebraRow}9 & Mug & 325 & Put the yellow-and-white mug in the microwave and close the door. \\
\midrule
\rowcolor{groupRow}\multicolumn{4}{@{}l}{\emph{LIBERO-Long-90 (part)}} \\
1  & Bowl atop & 175 & Close the top drawer and place the black bowl on top of the cabinet. \\
\rowcolor{zebraRow}3  & Rear butter & 171 & Put the rear butter in the top drawer and close it. \\
4  & Front butter & 172 & Put the front butter in the top drawer and close it. \\
\rowcolor{zebraRow}5  & Pudding & 159 & Put the chocolate pudding in the top drawer and close it. \\
8  & Bowl & 188 & Open the top drawer and put the bowl inside. \\
\rowcolor{zebraRow}21 & Pan (K.~3) & 262 & Turn on the stove and place the frying pan on it. \\
23 & Open/close & 210 & Close the bottom drawer and open the top drawer. \\
\rowcolor{zebraRow}45 & Pan (K.~9) & 227 & Turn on the stove and place the frying pan on it. \\
63 & Stack L$\rightarrow$R & 198 & Stack the left bowl on the right bowl and place the stack in the tray. \\
\rowcolor{zebraRow}64 & Stack R$\rightarrow$L & 216 & Stack the right bowl on the left bowl and place the stack in the tray. \\
\bottomrule
\end{tabularx}
\end{table}

\begin{table*}[t]
\centering
\small
\setlength{\tabcolsep}{5pt}
\definecolor{zebraRow}{RGB}{242,244,248}
\caption{RoboTwin-MeM task descriptions and benchmark statistics. The keyframe column reports the
number of required intermediate keyframes per episode. A range indicates
that the count varies across episodes. Mean steps reports the average over
successful trajectories.}
\label{tab:supp-robotwin-tasks}
\begin{tabularx}{\textwidth}{@{}l@{\hspace{8pt}}c@{\hspace{8pt}}c@{\hspace{8pt}}c@{\hspace{8pt}}>{\raggedright\arraybackslash}X@{}}
\toprule
Task & Episodes & \shortstack{Mean\\steps} &
\shortstack{Intermediate\\keyframes} & Task Description \\
\midrule
Cover Blocks Hard & 50 & 1442 & 4 &
Inspect and reclose each cover. Recall the revealed colors to reopen the corresponding covers in red, green, blue, and yellow order. \\
\rowcolor{zebraRow}Press Button Keyframe & 50 & 417 & 2--5 &
Read two cards to determine how many times to press the left and middle buttons, respectively. Finish with one press of the right button. \\
Put Back Block Hard & 50 & 1541 & 2 &
In each row, move the center block to a selected outer pad, return it to the center, and press the button. Then place both blocks on their previously visited pads and press again. \\
\rowcolor{zebraRow}Reproduce Route & 50 & 1338 & 4 &
Visit four pads in a random order with the center block, returning it to the center after each visit. Repeat the pad order with the outer block. \\
Pick Objects in Order & 50 & 1047 & 3 &
Inspect the objects under the covers, close the covers, and retrieve the objects in the order in which they were revealed. \\
\rowcolor{zebraRow}Pick the Unhidden Block & 50 & 695 & 3 &
Inspect and reclose the covers, then pick the exposed block whose color was absent under the covers. \\
Find Seal and Seal Stamp & 50 & 1205 & 1--4 &
Search the covered locations for the seal, use it to stamp, and return it to the location where it was found. \\
\rowcolor{zebraRow}Rearrange Blocks Hard & 50 & 837 & 1 &
Move one block to the center, return it to its mat, and then center the other block. Press the button after each move. \\
\bottomrule
\end{tabularx}
\end{table*}

\subsection{RoboTwin-MeM}
\label{sec:supp-robotwin}

RoboTwin-MeM contains longer-horizon manipulation tasks that require task-dependent numbers of intermediate visual records.
The evaluated tasks require between one and five keyframes.
The temporal-memory pathway uses the fixed overhead \texttt{cam\_high} camera.

We evaluate EventVLA with its native KEM selection and with TEEM using the same EventVLA checkpoint. 
We also report the previously available Official KEM results and the non-memory and memory-based comparison methods included in the benchmark. 

The ablation experiments use the same EventVLA checkpoint and evaluation conditions as the full TEEM experiment.
The \emph{w/o Target} setting replaces target-region quality with whole-image quality. 
The \emph{w/o Motion} setting removes motion-prominence-based candidate selection and selects a candidate after a fixed interval from the event-window opening. 
The \emph{w/o Stability} setting removes the temporal-stability ranking while preserving one evidence slot per plan node.

RoboTwin-MeM~\cite{yang2026eventvla} comprises eight simulation tasks in which later actions depend on information observed earlier. 
For example, in \emph{Reproduce Route}, the robot first visits four pads with one block and then repeats the same order with another block.
In \emph{Cover Blocks Hard}, it must remember which color was revealed under each cover after the covers have been closed. Table~\ref{tab:supp-robotwin-tasks} summarizes the eight tasks and their intermediate keyframe requirements.

\section{Physical-Robot Evaluation}
\label{sec:supp-real-world}

\subsection{Real-World Policy and Sampling Configurations}
\label{sec:realworld-config}

\paragraph{Policy fine-tuning.}
We fine-tune a $\pi_{0.5}$ policy with a PaliGemma vision-language backbone using LoRA. The policy receives images from a fixed Intel RealSense D435 camera, resized to $224\times224$, and text sequences of up to 200 tokens. It predicts action chunks of horizon $H=16$ using a 32-dimensional action representation. The CR5V controller consumes 7-dimensional actions, and the robot state is represented by 7 values. We train with AdamW ($\beta_1=0.9$, $\beta_2=0.95$, $\epsilon=10^{-8}$), weight decay $10^{-10}$, and a cosine learning-rate schedule from $2.5\times10^{-5}$ to $2.5\times10^{-6}$ after 50 warm-up steps. Training uses a global batch size of 32.

\begin{table}[!t]
\centering
\footnotesize
\setlength{\tabcolsep}{6pt}
\renewcommand{\arraystretch}{1.05}
\caption{Real-world $\pi_{0.5}$ policy fine-tuning configurations.}
\label{tab:realworld-config}
\begin{tabular}{@{}ll@{}}
\toprule
\multicolumn{2}{@{}l}{\itshape Model and inputs} \\
\midrule
Base VLM                   & PaliGemma ($\pi_{0.5}$) \\
Fine-tuning method         & LoRA \\
Camera                     & Fixed Intel RealSense D435 \\
Input resolution           & $224\times224$ \\
Max.\ text length          & 200 tokens \\
Action horizon $H$         & 16 \\
Action dim.\ (model/robot) & 32 / 7 \\
Robot state dim.           & 7 \\
\midrule
\multicolumn{2}{@{}l}{\itshape Optimization} \\
\midrule
Optimizer                  & AdamW \\
AdamW hyperparameters      & $\beta_1{=}0.9$, $\beta_2{=}0.95$, $\epsilon{=}10^{-8}$ \\
Weight decay               & $1\times10^{-10}$ \\
Learning rate              & $2.5\times10^{-5}\to2.5\times10^{-6}$ \\
LR schedule                & Cosine decay to the minimum \\
Warm-up steps              & 50 \\
Global batch size          & 32 \\
\bottomrule
\end{tabular}
\end{table}

\paragraph{TEEM at inference.}
During policy fine-tuning, the policy is provided with a causal Oracle memory trace: the initial record is available at the start, and each subsequent record is added only after its corresponding event. At real-robot inference, TEEM replaces this Oracle trace and selects records online from the D435 observations. We use a motion-prominence threshold of $P\geq0.40$, disable the $D$ gate, and use 50-step candidate chunks with soft commit and a 40-step occurrence lock.

We deploy TEEM on a DOBOT CR5V robot and evaluate three non-Markovian manipulation tasks. 
The tasks are \texttt{find\_block\_easy}, \texttt{pick\_x\_time}, and \texttt{find\_block\_hard}.
They require one, between two and four, and three keyframes, respectively.

We compare memory-free $\pi_{0.5}$, EventVLA fine-tuned with a $\pi_{0.5}$ backbone, and $\pi_{0.5}$ augmented with TEEM.
All methods use the same robot, task instructions, initial-state protocol, and success criteria.
The reported success rates are 0\%, 80\%, and 95\% for \texttt{find\_block\_easy}, 0\%, 90\%, and 100\% for \texttt{pick\_x\_time}, and 0\%, 65\%, and 85\% for \texttt{find\_block\_hard}, respectively.

\section{Qualitative Visualizations}
\label{sec:visualizations}

This section visualizes representative task executions and the intermediate visual information required by later actions. 
The examples show how a temporarily visible target state or an earlier interaction guides the subsequent steps of a long-horizon task.

\subsection{Failure Rollout Analysis}
\label{sec:failure-rollout-analysis}

\begin{figure}[t]
\centering
\begin{subfigure}{0.32\textwidth}
  \includegraphics[width=\linewidth]{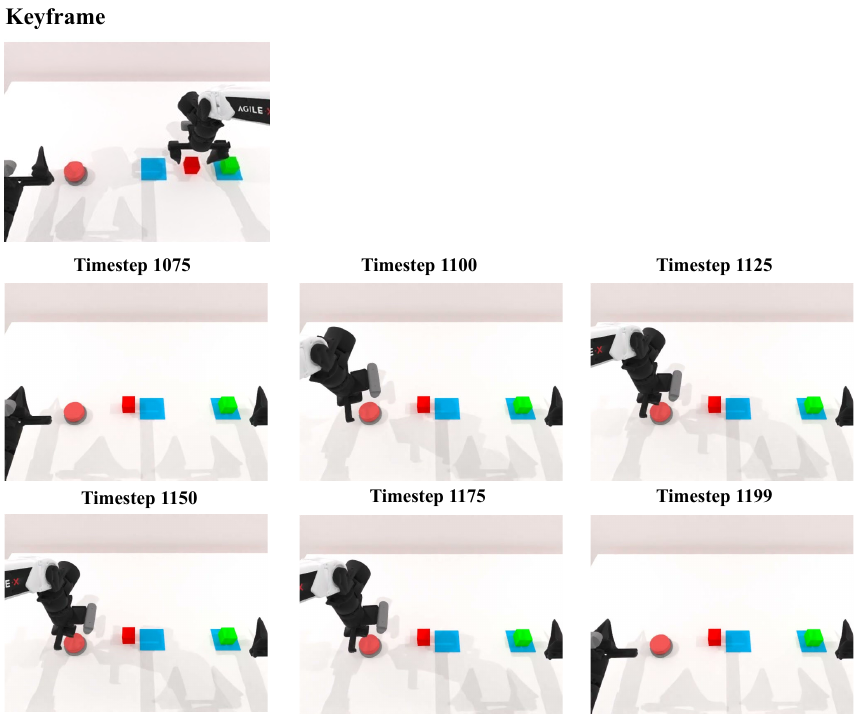}
  \caption{Rearrange Blocks Hard.}
\end{subfigure}\hfill
\begin{subfigure}{0.32\textwidth}
  \includegraphics[width=\linewidth]{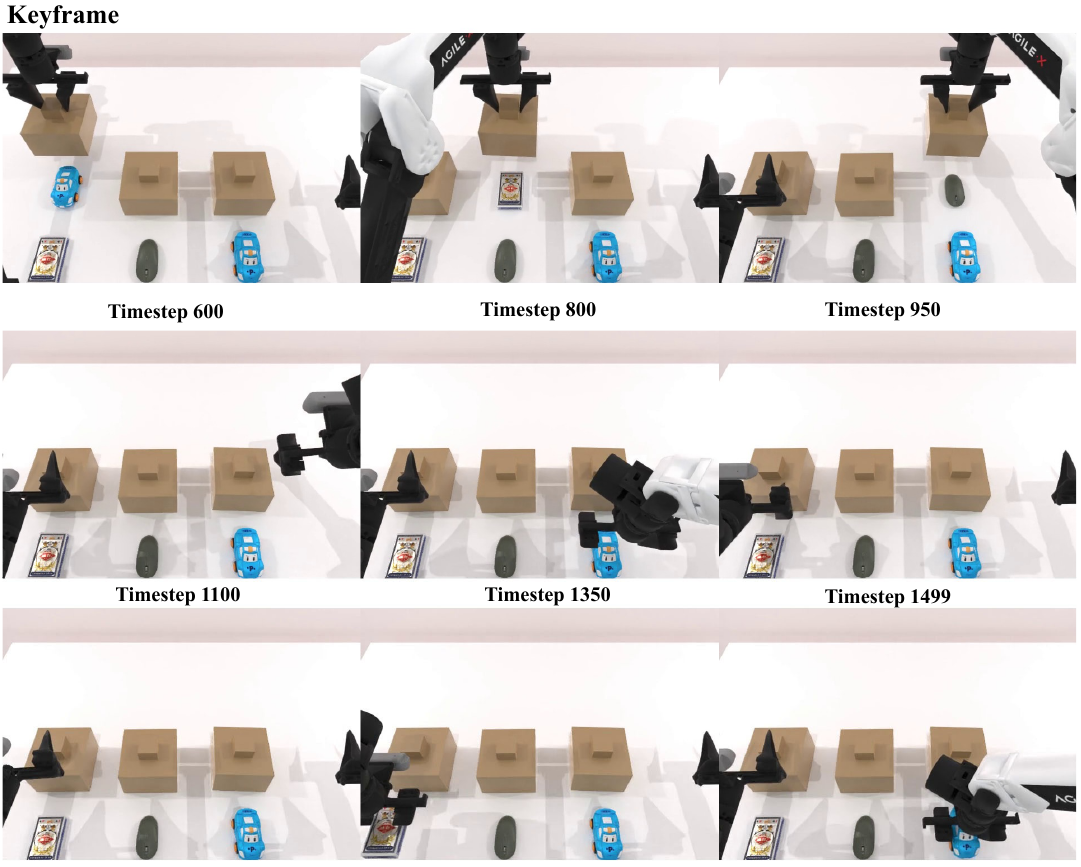}
  \caption{Pick Objects in Order.}
\end{subfigure}\hfill
\begin{subfigure}{0.32\textwidth}
  \includegraphics[width=\linewidth]{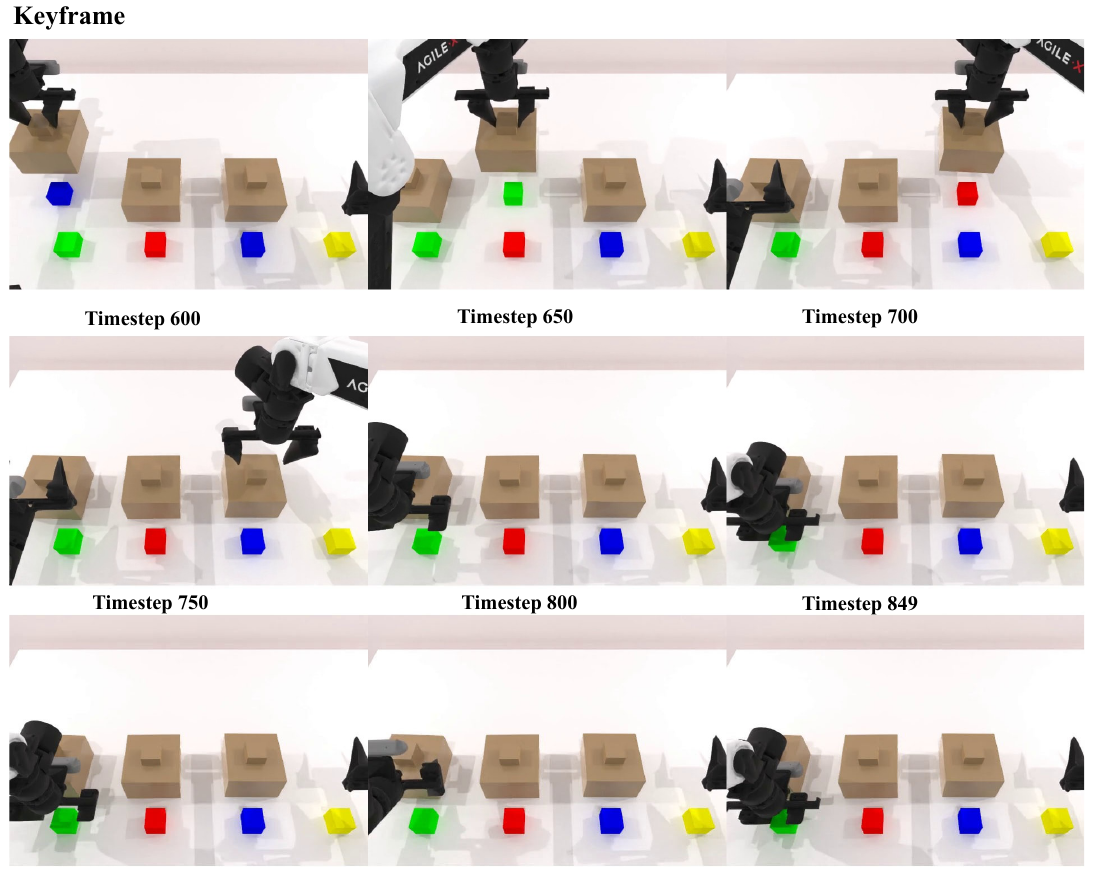}
  \caption{Pick the Unhidden Block.}
\end{subfigure}
\caption{\textbf{Failure cases.} (a) TEEM's evidence from source step 129 is
delivered at query step 150; the rollout reaches a maximum reward of 0.80 but
the green block stays on its original pad. (b) A maximum reward of 0.50 is
reached, yet the blue car remains on the table at the 1500-step limit.
(c) Three frames recording the cover-inspection results identify the yellow
block as the target, but during steps 700--849 the gripper repeatedly approaches
the green-block region.}
\label{fig:fail-cases}
\end{figure}

\begin{figure*}[h]
  \centering
  \includegraphics[width=0.85\textwidth]{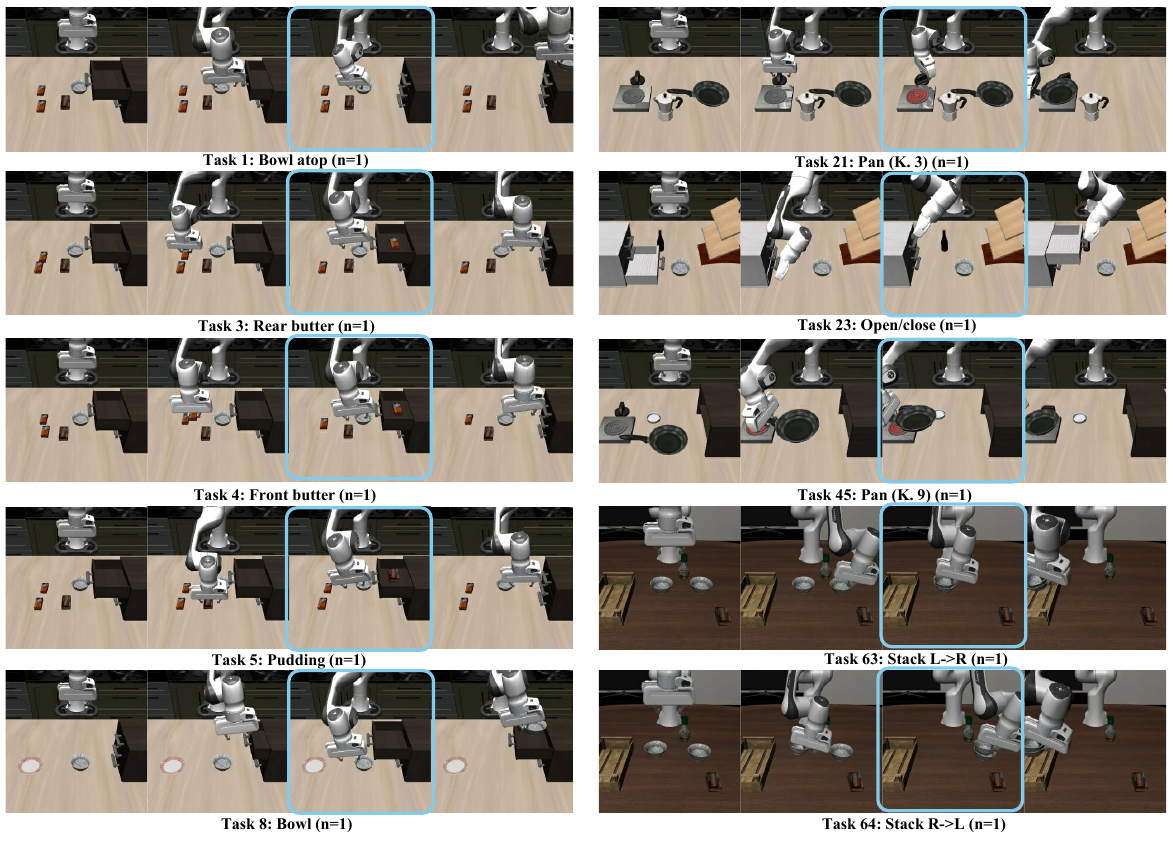}
  \caption{Representative MemoryVLA + TEEM rollouts on selected LIBERO-90 tasks.
  Each sequence shows the initial scene, intermediate task states, the TEEM-selected evidence frame(s) highlighted in orange, and the final state.}
  \label{fig:libero90-rollouts}
\end{figure*}

Firstly, we present and analyze failed RoboTwin-MeM rollouts, examining how task execution can fail even after TEEM has captured and delivered the relevant evidence to the policy.
Across these examples, the visualizations show that evidence acquisition and successful downstream interaction are distinct stages.

In \emph{Rearrange Blocks Hard} (seed 100004), the rollout reaches a maximum reward of 0.80.
According to the task state machine, the first block has been moved to the center and returned to its original pad, with the first two button presses completed. 
The remaining sequence requires moving the second block to the center and pressing the button for the third time. 
TEEM's record from source step 129 is delivered to the policy at query step 150.
In the final state, the red block is on the center pad, while the green block remains on its original pad. 
The video shows repeated approaches to the green block, but no completed transfer. 
The failure therefore occurs during the final block-moving stage, after the earlier evidence has been delivered.

In \emph{Pick Objects in Order} (seed 100000), the episode reaches the 1500-step limit with a maximum reward of 0.50.
This reward corresponds to progress through the cover-inspection stage, but not to successful object pickup. 
The visualization shows the delivered memory frame followed by repeated policy interactions near the blue car; the car remains on the table at termination.

In \emph{Pick the Unhidden Block} (seed 100000), three memory frames record the outcomes of the cover inspections, indicating that the target block is yellow. 
However, during steps 700--849, the gripper repeatedly approaches the green-block region, and the episode terminates with a reward of 0.45 without completing the pickup. 
This rollout shows that the inspection evidence was captured and delivered, while the subsequent target interaction did not complete successfully.

Together, these cases illustrate that capturing task-relevant keyframes does not by itself ensure successful execution: the policy must also use the retained evidence to select and manipulate the appropriate object in later stages.
This is a limitation of the native model's capability. 
Currently, TEEM only handles accurately identifying evidence based on the target; in future work, we will investigate how to further improve task success rates.

\subsection{Simulation Rollouts in LIBERO-90}
\label{sec:visualizations-libero}

Figure~\ref{fig:libero90-rollouts} shows representative MemoryVLA + TEEM rollouts from the selected LIBERO-90 subset. 
Each sequence follows the episode from its initial state through a TEEM-selected observation to task completion. 
The highlighted frame marks the visual evidence retained for a later step, while the surrounding frames show the action context in which that evidence is acquired and used.

For example, task 64 requires stacking the right bowl on the left bowl and then placing the stack in the tray. 
The sequence first establishes the bowl configuration, then shows the stack formed by the required ordering, and finally shows the completed placement in the tray. 
This example illustrates why the intermediate bowl arrangement matters to the later placement step. 
A second example, task 21, requires turning on the stove before placing the pan on it. 
Its sequence shows the stove state and the subsequent pan placement, illustrating how an earlier scene change can remain relevant to the following action.

\subsection{Real-World Robot Execution Sequences}
\label{sec:visualizations-real-world}

\begin{figure*}[h]
  \centering
  \includegraphics[width=0.85\textwidth]{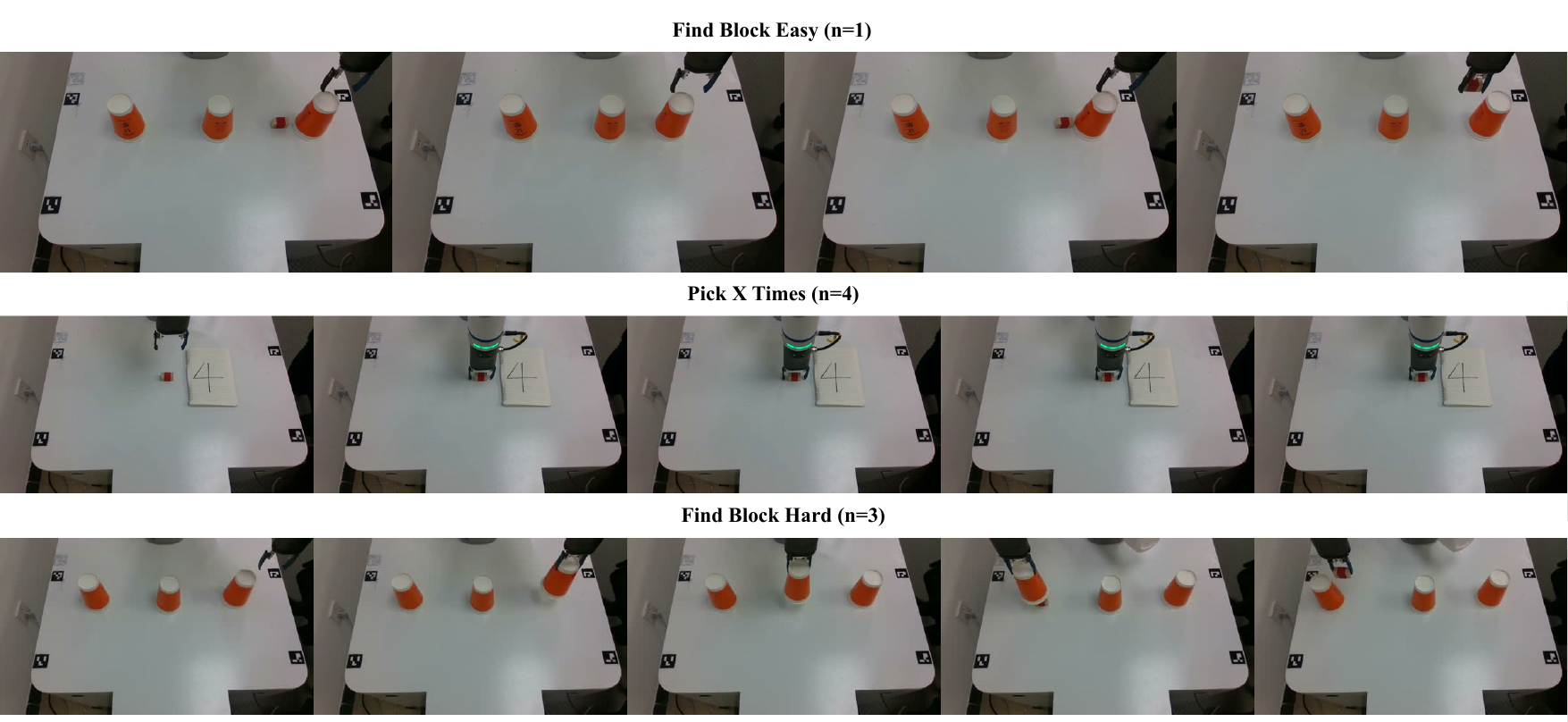}
  \caption{Representative TEEM-enabled task sequences on the DOBOT CR5V.
  The rows show \texttt{find\_block\_easy},
  \texttt{pick\_x\_time}, and \texttt{find\_block\_hard}, respectively.
  Orange outlines indicate observations retained by TEEM. Each sequence
  follows the task from the initial scene through evidence acquisition to
  the subsequent manipulation steps.}
  \label{fig:real-world-sequences}
\end{figure*}

Figure~\ref{fig:real-world-sequences} shows representative executions on the DOBOT CR5V. 
The three tasks require the robot to retain different kinds of information across steps.
In \texttt{find\_block\_easy}, the robot first covers the block with the nearest cup, lifts and moves that cup aside, and then picks up the exposed block.
In \texttt{pick\_x\_time}, it reads the count shown on the paper and repeats the block pick-and-place cycle the specified number of times. 
In \texttt{find\_block\_hard}, it inspects the cups from left to right, identifies the cup containing the block, reopens that cup, and picks up the block.

The sequences show the selected visual records alongside the subsequent robot actions. 
In \texttt{find\_block\_easy} and \texttt{find\_block\_hard}, the retained observation records which location contains the block. 
In \texttt{pick\_x\_time}, successive records preserve the paper count and the interaction progress needed to complete the requested number of cycles.
Together, these examples illustrate how task-relevant observations acquired at different stages support later actions on the physical robot.

\section{Limitations and Future Work}
\label{sec:supp-limitations}

On RoboTwin-MeM, TEEM still requires task-specific adjustments to the task-family specification to accommodate EventVLA's native memory interface. These adjustments define how the task's multiple keyframes map to ordered subtasks and target regions, while leaving the policy weights unchanged.
By contrast, for the evaluated LIBERO-Long tasks, the VLM-generated plan can be used directly without further task-specific adjustment. 
We will investigate more general methods for identifying subtask boundaries and assigning their targets, reducing the need for manual configuration.
On shorter-horizon LIBERO-Long tasks with visually cluttered scenes, the target-region quality score $Q$ may not always identify the most informative frame. Future work could incorporate lightweight semantic assessment to improve keyframe selection in these settings.
Our analysis of RoboTwin-MeM failures shows that capturing all required keyframes does not guarantee successful downstream execution. 
The policy may still repeat grasp attempts or fail in a later action. Future work could connect evidence selection with memory retrieval and action outcomes to develop a closed-loop evidence-acquisition process.
Finally, our evaluation covers selected tasks from LIBERO-Long and RoboTwin-MeM, as well as three physical-robot tasks. 
Evaluations across additional task families, robot platforms, camera configurations, and policy-native memory mechanisms will help establish TEEM's broader applicability.

\end{document}